\documentclass[10pt,twocolumn,letterpaper]{article}

\usepackage[pagenumbers]{cvpr} 

\usepackage{array,multirow,pifont}
\usepackage[dvipsnames]{xcolor}

\renewcommand{\paragraph}[1]{\vspace{1.25mm}\noindent\textbf{#1}}
\newlength\savewidth\newcommand\shline{\noalign{\global\savewidth\arrayrulewidth
  \global\arrayrulewidth 1pt}\hline\noalign{\global\arrayrulewidth\savewidth}}
\newcommand\hshline{\noalign{\global\savewidth\arrayrulewidth
  \global\arrayrulewidth 0.5pt}\hline\noalign{\global\arrayrulewidth\savewidth}}
\newcommand{\tablestyle}[2]{\setlength{\tabcolsep}{#1}\renewcommand{\arraystretch}{#2}\centering\footnotesize}
\newcolumntype{x}[1]{>{\centering\arraybackslash}p{#1pt}}
\newcolumntype{y}[1]{>{\raggedright\arraybackslash}p{#1pt}}
\newcolumntype{z}[1]{>{\raggedleft\arraybackslash}p{#1pt}}
\definecolor{degray}{gray}{.6}
\newcommand{\deemph}[1]{\textcolor{degray}{#1}}
\def\x0{z_0}
\def\xp0{z'}
\def\xt{z_t}
\def\xtm1{z_{t-1}}
\def\epst{\epsilon}

\def\alphat{\gamma_t}
\def\sigmat{\sigma_t}

\newcommand{\app}{\raise.17ex\hbox{$\scriptstyle\sim$}}
\def\ours{\emph{l}-DAE\xspace}
\definecolor{cvprblue}{rgb}{0.21,0.49,0.74}
\usepackage[pagebackref=false,breaklinks,colorlinks,citecolor=cvprblue]{hyperref}
\usepackage[british, english, american]{babel}

\def\paperID{8751} 
\def\confName{CVPR}
\def\confYear{2024}

\title{\Large Deconstructing Denoising Diffusion Models for Self-Supervised Learning}

\author{Xinlei Chen\textsuperscript{1} \qquad
Zhuang Liu\textsuperscript{1} \qquad
Saining Xie\textsuperscript{2} \qquad
Kaiming He\textsuperscript{1} \\[3mm]
\textsuperscript{1}FAIR, Meta \qquad \textsuperscript{2}New York University\vspace{-2mm}}

\begin{document}
\maketitle
\begin{abstract}
\vspace{-.5em}
In this study, we examine the representation learning abilities of Denoising Diffusion Models (DDM) that were originally purposed for image generation. Our philosophy is to \emph{deconstruct} a DDM, gradually transforming it into a classical Denoising Autoencoder (DAE). This deconstructive procedure allows us to explore how various components of modern DDMs influence self-supervised representation learning. We observe that only a very few modern components are critical for learning good representations, while many others are nonessential. Our study ultimately arrives at an approach that is highly simplified and to a large extent resembles a classical DAE. We hope our study will rekindle interest in a family of classical methods within the realm of modern self-supervised learning. 
\end{abstract}

\section{Introduction\label{sec:intro}}

Denoising is at the core in the current trend of generative models in computer vision and other areas. Popularly known as \textit{Denoising Diffusion Models} (DDM) today, these methods~\cite{SohlDickstein2015,Song2019,Song2020,Ho2020,Nichol2021,Dhariwal2021} learn a \textit{Denoising Autoencoder} (DAE) \cite{Vincent2008} that removes noise of multiple levels driven by a diffusion process. These methods achieve impressive image generation quality, especially for high-resolution, photo-realistic images~\cite{Rombach2022,Peebles2023}---in fact, these \textit{generation} models are so good that they appear to have strong \textit{recognition} representations for understanding the visual content.

\begin{figure}[t]
\centering
\begin{center}
\includegraphics[width=.99\linewidth]{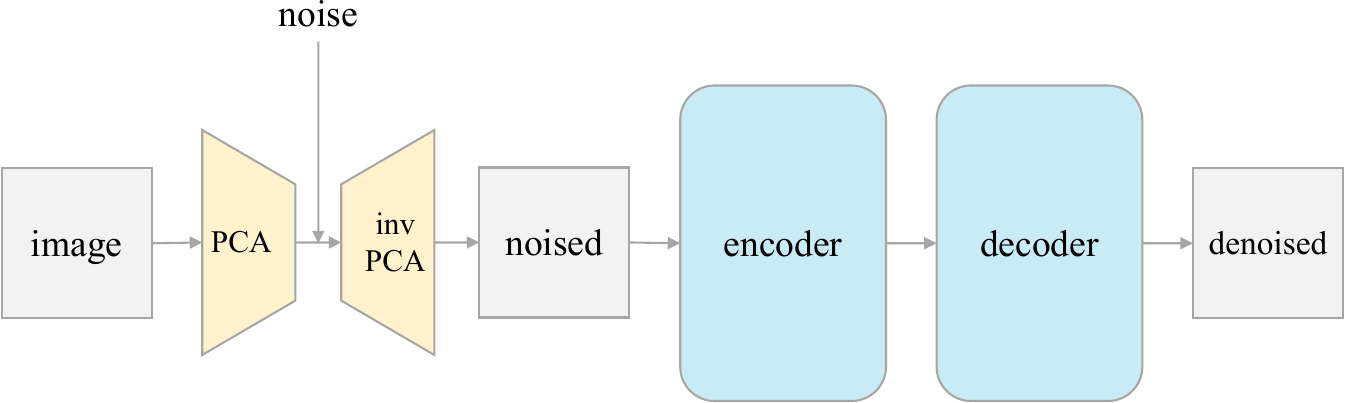}
\end{center}
\vspace{-.1em}
\scriptsize{
\begin{tabular}{c@{\hskip .2mm}c@{\hskip .2mm}c}
\includegraphics[width=.31\linewidth]{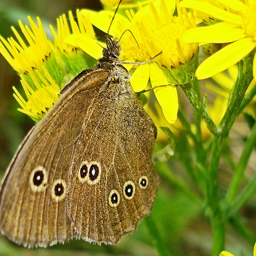} & \includegraphics[width=.31\linewidth]{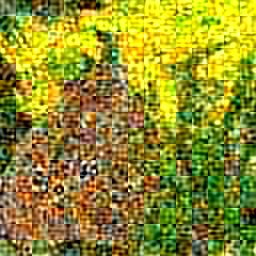} & \includegraphics[width=.31\linewidth]{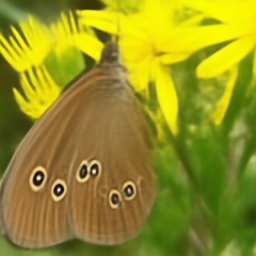} \\
clean image & noised & denoised
\end{tabular}
}
\vspace{-.5em}
\caption{\label{fig:teaser}
The {latent Denoising Autoencoder (\textbf{\ours})} architecture we have ultimately reached, after a thorough exploration of {deconstructing} Denoising Diffusion Models (DDM)~\cite{Ho2020}, with the goal of approaching the classical Denoising Autoencoder (DAE)~\cite{Vincent2008} as much as possible. Here, the clean image (left) is projected onto a latent space using patch-wise PCA, in which noise is added (middle). It is then projected back to pixels via inverse PCA.
An autoencoder is learned to predict a denoised image (right). This simple architecture largely resembles classical DAE (with the main difference that noise is added to the latent) and achieves competitive self-supervised learning performance.
}
\vspace{-.5em}
\end{figure}

While DAE is a powerhouse of today's generative models, it was originally proposed for learning representations \cite{Vincent2008} from data in a self-supervised manner. In today's community of representation learning, the arguably most successful variants of DAEs are based on ``\textit{masking noise}" \cite{Vincent2008}, such as predicting missing text in languages (\eg, BERT~\cite{Devlin2019}) or missing patches in images (\eg, MAE~\cite{He2022}). However, in concept, these masking-based variants remain significantly different from removing additive (\eg, Gaussian) noise: while the masked tokens explicitly specify unknown \vs known content, no clean signal is available in the task of separating additive noise. Nevertheless, today's DDMs for generation are dominantly based on additive noise, implying that they may learn representations without explicitly marking unknown/known content.

Most recently, there has been an increasing interest \cite{Xiang2023,Mukhopadhyay2023} in inspecting the representation learning ability of DDMs. In particular, these studies directly take \textit{off-the-shelf} pre-trained DDMs~\cite{Ho2020,Peebles2023,Dhariwal2021}, which are originally purposed for generation, and evaluate their representation quality for recognition. They report encouraging results using these \textit{generation-oriented} models. However, these pioneering studies obviously leave open questions: these off-the-shelf models were designed for generation, not recognition; it remains largely unclear whether the representation capability is gained by a denoising-driven process, or a diffusion-driven process.

In this work, we take a much deeper dive into the direction initialized by these recent explorations \cite{Xiang2023,Mukhopadhyay2023}. Instead of using an off-the-shelf DDM that is generation-oriented, we train models that are recognition-oriented. At the core of our philosophy is to \textit{deconstruct} a DDM, changing it step-by-step into a classical DAE. Through this \mbox{deconstructive} research process, we examine every single aspect (that we can think of) of a modern DDM, with the goal of learning representations. This research process gains us new understandings on what are the critical components for a DAE to learn good representations.

Surprisingly, we discover that the main critical component is a tokenizer \cite{Rombach2022} that creates a \textit{low-dimensional latent} space. Interestingly, this observation is largely \textit{independent} of the specifics of the tokenizer---we explore a standard VAE \cite{Kingma2013}, a patch-wise VAE, a patch-wise AE, and a patch-wise PCA encoder. We discover that it is the \textit{low-dimensional} latent space, rather than the tokenizer specifics, that enables a DAE to achieve good representations.

Thanks to the effectiveness of PCA, our deconstructive trajectory ultimately reaches a simple architecture that is highly similar to the classical DAE (\cref{fig:teaser}). We project the image onto a latent space using patch-wise PCA, add noise, and then project it back by inverse PCA. Then we train an autoencoder to predict a denoised image. We call this architecture ``latent Denoising Autoencoder" (\ours).

Our deconstructive trajectory also reveals many other intriguing properties that lie between DDM and classical DAE. For one example, we discover that even using \textit{a single noise level} (\ie, not using the noise scheduling of DDM) can achieve a decent result with our \ours. The role of using multiple levels of noise is analogous to a form of data augmentation, which can be beneficial, but not an enabling factor. With this and other observations, we argue that the representation capability of DDM is mainly gained by the denoising-driven process, not a diffusion-driven process.

Finally, we compare our results with previous baselines. On one hand, our results are substantially better than the off-the-shelf counterparts (following the spirit of \cite{Xiang2023,Mukhopadhyay2023}): this is as expected, because these are our starting point of deconstruction. On the other hand, our results fall short of baseline contrastive learning methods (\eg, \cite{Chen2021a}) and masking-based methods (\eg, \cite{He2022}), but the gap is reduced. Our study suggests more room for further research along the direction of DAE and DDM. 

\section{Related Work\label{sec:related}}

In the history of machine learning and computer vision, the generation of images (or other content) has been closely intertwined with the development of unsupervised or self-supervised learning. 
Approaches in generation are conceptually forms of un-/self-supervised learning, where models were trained without labeled data, learning to capture the underlying distributions of the input data.

There has been a prevailing belief that the ability of a model to generate high-fidelity data is indicative of its potential for learning good representations. Generative Adversarial Networks (GAN)~\cite{Goodfellow2014}, for example, have ignited broad interest in adversarial representation learning \cite{Donahue2017,Donahue2019}.
Variational Autoencoders (VAEs)~\cite{Kingma2013}, originally conceptualized as generative models for approximating data distributions, have evolved to become a standard in learning {localized representations} (``tokens"), \eg, VQVAE \cite{Oord2017} and variants \cite{Esser2021}. Image inpainting \cite{Bertalmio2000}, essentially a form of conditional image generation, has led to a family of modern representation learning methods, including Context Encoder \cite{Pathak2016} and Masked Autoencoder (MAE) \cite{He2022}.

Analogously, the outstanding generative performance of Denoising Diffusion Models (DDM) \cite{SohlDickstein2015,Song2019,Song2020,Ho2020,Dhariwal2021} has drawn attention for their potential in representation learning. Pioneering studies \cite{Xiang2023,Mukhopadhyay2023} have begun to investigate this direction by evaluating existing pre-trained DDM networks. However, we note that while a model's generation capability suggests a certain level of understanding, it does not necessarily translate to representations useful for downstream tasks. Our study delves deeper into these issues.

On the other hand, although Denoising Autoencoders (DAE) \cite{Vincent2008} have laid the groundwork for autoencoding-based representation learning, their success has been mainly confined to scenarios involving masking-based corruption (\eg,~\cite{He2022,Xie2022,Fang2022,Chen2023}). To the best of our knowledge, little or no recent research has reported results on classical DAE variants with additive Gaussian noise, and we believe that the underlying reason is that a simple DAE baseline (\cref{fig:dit_arch}(a)) performs poorly\footnotemark~(\eg, in $\app$20\% \cref{fig:pca_pixel}). 

\footnotetext{According to the authors of MoCo \cite{He2020} and MAE \cite{He2022}, significant effort has been devoted to DAE baselines during the development of those works, following the best practice established. However, it has not led to meaningful results ($<$20\% accuracy).}

\section{Background: Denoising Diffusion Models\label{sec:ddpm}}

Our deconstructive research starts with a Denoising Diffusion Model (DDM)~\cite{SohlDickstein2015,Song2019,Song2020,Ho2020,Dhariwal2021}. We briefly describe the DDM we use, following \cite{Dhariwal2021,Peebles2023}.

A diffusion process starts from a clean data point ($\x0$) and sequentially adds noise to it. At a specified time step $t$, the noised data $\xt$ is given by:
\begin{equation}\label{eq:diffuse}
    \xt = \alphat\x0 + \sigmat\epst
\end{equation}
where $\epst{\sim}\mathcal{N}(0, \mathbf{I})$ is a noise map sampled from a Gaussian distribution, and $\alphat$ and $\sigmat$ define the scaling factors of the signal and of the noise, respectively. By default, it is set $\alphat^2+\sigmat^2=1$ \cite{Nichol2021,Dhariwal2021}.

\begin{figure}[t]
\vspace{-1em}
\centering
\begin{subfigure}[t]{\linewidth}
    \centering
    \includegraphics[width=.98\linewidth, trim=0 0 0 75mm, clip]{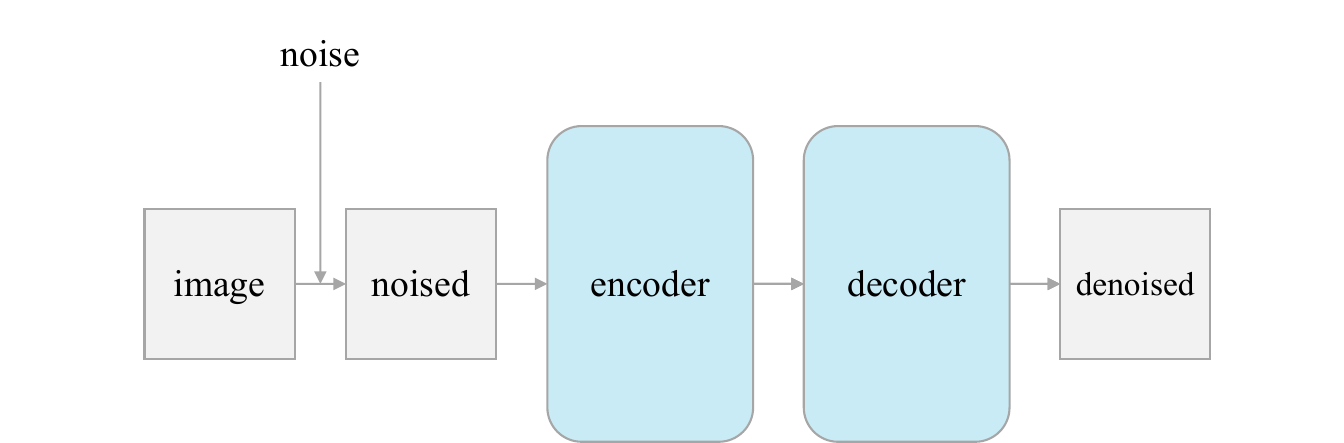}
    \vspace{.2em}
    \caption{a classical \textbf{Denoising Autoencoders} (DAE)}
\end{subfigure}
\\
\vspace{1em}
\begin{subfigure}[t]{\linewidth}
    \centering
    \includegraphics[width=.98\linewidth, trim=0 75mm 0 0, clip]{fig/dit-arch.pdf}
    \caption{a modern \textbf{Denoising Diffusion Model} (DDM) on a latent space}
\end{subfigure}
\vspace{-.1em}
\caption{\label{fig:dit_arch}\textbf{A classical DAE and a modern DDM}. 
(a) A classical DAE that adds and predicts noise on the image space.
(b) State-of-the-art DDMs (\eg, LDM~\cite{Rombach2022}, DIT~\cite{Peebles2023}) that operate on a latent space, where the noise is added and predicted.
}
\end{figure}

A denoising diffusion model is learned to remove the noise, conditioned on the time step $t$. 
Unlike the original DAE \cite{Vincent2008} that predicts a clean input, 
the modern DDM~\cite{Ho2020,Nichol2021} often predicts the noise $\epst$. Specifically, a loss function in this form is minimized:
\begin{equation}\label{eq:loss}
    \|\epst - \mathtt{net}(z_t)\|^2
\end{equation}
where $\mathtt{net}(z_t)$ is the network output.
The network is trained for multiple noise levels given a noise \emph{schedule}, conditioned on the time step $t$.
In the generation process, a trained model is iteratively applied until it reaches the clean signal $\x0$.

DDMs can operate on two types of input spaces.
One is the original pixel space~\cite{Dhariwal2021}, where the raw image $x_0$ is directly used as $\x0$. The other option is to build DDMs on a \emph{latent} space produced by a \textit{tokenizer}, following~\cite{Rombach2022}. See \cref{fig:dit_arch}(b).
In this case, a pre-trained tokenizer $f$ (which is often another autoencoder, \eg, VQVAE~\cite{Oord2017}) is used to map the image $x_0$ into its latent $\x0{=}f(x_0)$. 

\paragraph{Diffusion Transformer (DiT).} 
Our study begins with the Diffusion Transformer (DiT)~\cite{Peebles2023}. We choose this Transformer-based DDM for several reasons: (i) Unlike other UNet-based DDMs~\cite{Dhariwal2021,Rombach2022}, Transformer-based architectures can provide fairer comparisons with other self-supervised learning baselines driven by Transformers (\eg, \cite{Chen2021a,He2022}); 
(ii) DiT has a clearer distinction between the encoder and decoder, while a UNet's encoder and decoder are connected by skip connections and may require extra effort on network surgery when evaluating the encoder;
(iii) DiT trains much faster than other UNet-based DDMs (see~\cite{Peebles2023}) while achieving better generation quality.

We use the DiT-Large (\textbf{DiT-L}) variant~\cite{Peebles2023} as our DDM baseline. 
In DiT-L, the encoder and decoder \textit{put together} have the size of ViT-L \cite{Dosovitskiy2021} (24 blocks). We evaluate the representation quality (linear probe accuracy) of the \textit{encoder}, which has 12 blocks, referred to as ``$\frac{1}{2}$L" (half large).

\paragraph{Tokenizer.} 
DiT instantiated in~\cite{Peebles2023} is a form of Latent Diffusion Models (LDM)~\cite{Rombach2022}, which uses a VQGAN tokenizer~\cite{Esser2021}. Specifically, this VQGAN tokenizer transforms the 256$\times$256$\times$3 input image (height$\times$width$\times$channels) into a 32$\times$32$\times$4 latent map, with a stride of 8.


\paragraph{Starting baseline.} 
By default, we train the models for 400 epochs on ImageNet~\cite{Deng2009} with a resolution of 256$\times$256 pixels.
Implementation details are in \cref{sec:impl}.

Our DiT baseline results are reported in \cref{tab:dit_ssl} (line 1).
With DiT-L, we report a linear probe accuracy of \textbf{57.5}\% using its $\frac{1}{2}$L encoder.
The generation quality (Fr\'{e}chet Inception Distance~\cite{Heusel2017}, FID-50K) of this DiT-L model is \textbf{11.6}. This is the starting point of our destructive trajectory.

Despite differences in implementation details, our starting point conceptually follows recent studies~\cite{Xiang2023,Mukhopadhyay2023} (more specifically, DDAE~\cite{Xiang2023}), which evaluate off-the-shelf DDMs under the linear probing protocol.

\section{Deconstructing Denoising Diffusion Models\label{sec:deconstruct}}

Our deconstruction trajectory is divided into three stages.
We first adapt the generation-focused settings in DiT to be more oriented toward self-supervised learning (\cref{sec:dit_ssl}).
Next, we deconstruct and simplify the tokenizer step by step (\cref{sec:decon_token}).
Finally, we attempt to reverse as many DDM-motivated designs as possible, pushing the models towards a classical DAE~\cite{Vincent2008} (\cref{sec:to_pixel}).
We summarize our learnings from this deconstructing process in \cref{sec:summary}.

\subsection{Reorienting DDM for Self-supervised Learning\label{sec:dit_ssl}}

While a DDM is conceptually a form of a DAE, it was originally developed for the purpose of image generation. Many designs in a DDM are oriented toward the generation task.
Some designs are \textit{not legitimate} for self-supervised learning (\eg, class labels are involved); some others are not necessary if visual quality is not concerned. In this subsection, we reorient our DDM baseline for the purpose of self-supervised learning, summarized in \cref{tab:dit_ssl}.


\paragraph{Remove class-conditioning.}
A high-quality DDM is often trained with conditioning on \textit{class labels}, which can largely improve the generation quality. But the usage of class labels is simply not legitimate in the context of our self-supervised learning study. As the first step, we remove class-conditioning in our baseline. 

Surprisingly, removing class-conditioning substantially improves the linear probe accuracy from 57.5\% to 62.1\% (\cref{tab:dit_ssl}), even though the generation quality is greatly hurt as expected (FID from 11.6 to 34.2).  
We hypothesize that directly conditioning the model on class labels may reduce the model's demands on encoding the information related to class labels. Removing the class-conditioning can force the model to learn more semantics.

\begin{table}[t]
\vspace{-1em}
\centering
\tablestyle{7pt}{1.1}
\begin{tabular}{y{130}x{30}z{30}}
 & acc. ($\uparrow$) & FID ($\downarrow$) \\
\shline
{DiT baseline} & \deemph{57.5} & {11.6}~ \\
\hline
{+ remove class-conditioning} & \deemph{62.5} & {30.9} \\
~~ + remove VQGAN perceptual loss & 58.4 & 54.3 \\
~~~~ + remove VQGAN adversarial loss & 59.0 & 75.6 \\
~~~~~~ + replace noise schedule & 63.4 & 93.2 \\
\end{tabular}

\vspace{-0.8em}
\caption{\label{tab:dit_ssl}\textbf{Reorienting DDM for self-supervised learning}.
We begin with the DiT~\cite{Peebles2023} baseline and evaluate its linear probe accuracy (acc.) on ImageNet.
Each line is based on a modification of the immediately preceding line. 
The entries in \deemph{gray}, in which class labels are used, are not legitimate results for self-supervised learning.
See \cref{sec:dit_ssl} for description.
}
\vspace{-1em}
\end{table}

\paragraph{Deconstruct VQGAN.} In our baseline, the VQGAN tokenizer, presented by LDM \cite{Rombach2022} and inherited by DiT, is trained with multiple loss terms: (i) autoencoding reconstruction loss; (ii) KL-divergence regularization loss \cite{Rombach2022};\footnotemark~(iii) perceptual loss~\cite{Zhang2018b} based on a \textit{supervised} VGG net~\cite{Simonyan2015} trained for ImageNet classification; and (iv) adversarial loss~\cite{Goodfellow2014,Esser2021} with a discriminator. We ablate the latter two terms in \cref{tab:dit_ssl}.

\footnotetext{The KL form in \cite{Rombach2022} does not perform explicit  vector quantization (VQ), interpreted as ``the quantization layer absorbed by the decoder" \cite{Rombach2022}.}

As the \textbf{perceptual loss}~\cite{Zhang2018b} involves a supervised pre-trained network, using the VQGAN trained with this loss is not legitimate. Instead, we train another VQGAN tokenizer \cite{Rombach2022} in which we remove the perceptual loss. Using this tokenizer \textit{reduces} the linear probe accuracy significantly from 62.5\% to 58.4\% (\cref{tab:dit_ssl}), which, however, provides the first legitimate entry thus far. This comparison reveals that \textit{a tokenizer trained with the perceptual loss (with class labels) in itself provides semantic representations}. We note that the perceptual loss is not used from now on, in the remaining part of this paper.

We train the next VQGAN tokenizer that further removes the \textbf{adversarial loss}. It slightly increases the linear probe accuracy from 58.4\% to 59.0\% (\cref{tab:dit_ssl}).
With this, our tokenizer at this point is essentially a VAE, which we move on to deconstruct in the next subsection. We also note that removing either loss harms generation quality.

\begin{figure}[t]
\vspace{-1em}
\centering
\includegraphics[width=.9\linewidth]{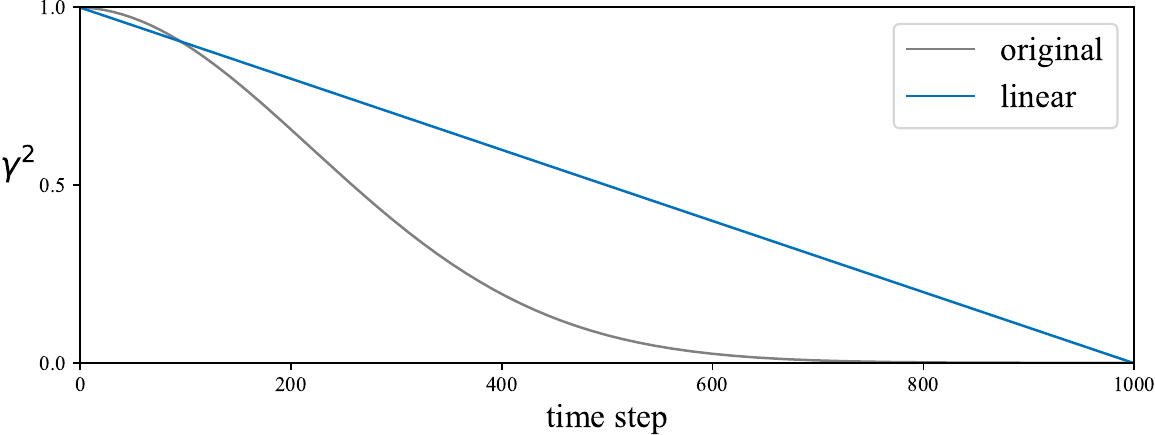}
\vspace{-.5em}
\caption{\label{fig:noise_sched}
\textbf{Noise schedules}.
The original schedule~\cite{Ho2020,Peebles2023}, which sets $\alphat^2{=}{\Pi_{s=1}^{t}{(1-\beta_s)}}$ with a linear schedule of $\beta$, spends many time steps on very noisy images (small $\gamma$). Instead, we use a simple schedule that is linear on $\gamma^2$, which provides less noisy images.
}
\vspace{-.5em}
\end{figure}

\paragraph{Replace noise schedule.} In the task of generation, the goal is to progressively turn a noise map into an image. As a result, the original noise schedule spends many time steps on very noisy images (\cref{fig:noise_sched}). This is not necessary if our model is not generation-oriented.

We study a simpler noise schedule for the purpose of self-supervised learning. Specifically, we let $\alphat^2$ linearly decay in the range of $1 {>} \alphat^2 {\geq} 0$ (\cref{fig:noise_sched}). This allows the model to spend more capacity on cleaner images.
This change greatly improves the linear probe accuracy from 59.0\% to 63.4\% (\cref{tab:dit_ssl}), suggesting that the original schedule focuses too much on noisier regimes. On the other hand, as expected, doing so further hurts the generation ability, leading to a FID of 93.2. 

\paragraph{Summary.}
Overall, the results in \cref{tab:dit_ssl} reveal that \textit{\textbf{self-supervised learning  performance is not correlated to generation quality}}. The representation capability of a DDM is not necessarily the outcome of its generation capability.

\subsection{Deconstructing the Tokenizer\label{sec:decon_token}}

Next, we further deconstruct the VAE tokenizer by making substantial simplifications. 
We compare the following four variants of autoencoders as the tokenizers, each of which is a simplified version of the preceding one:

\begin{itemize}
\item \paragraph{Convolutional VAE.} 
Our deconstruction thus far leads us to a VAE tokenizer. As common practice \cite{Kingma2013,Rombach2022}, the encoder $f(\cdot)$ and decoder $g(\cdot)$ of this VAE are deep convolutional (conv) neural networks \cite{LeCun1989}. This {convolutional VAE} minimizes the following loss function:
\begin{equation}
\|x - g(f(x))\|^2 + \mathbb{KL}\left[f(x)|\mathcal{N}\right]. \nonumber
\end{equation}
Here, $x$ is the input image of the VAE. The first term is the reconstruction loss, and the second term is the Kullback-Leibler divergence~\cite{Bishop2006,Esser2021} between the latent distribution of $f(x)$ and a unit Gaussian distribution.

\item \paragraph{Patch-wise VAE.} 
Next we consider a simplified case in which the VAE encoder and decoder are both \textit{linear} projections, and the VAE input $x$ is a \textit{patch}. 
The training process of this \textit{patch-wise VAE} minimizes this loss:
\begin{equation}
\|x - U{^T}Vx\|^2 + \mathbb{KL}\left[{{V}{x}}|\mathcal{N}\right]. \nonumber
\end{equation}
Here $x$ denotes a patch flattened into a $D$-dimensional vector. Both $U$ and $V$ are $d{\times}{D}$ matrixes, where $d$ is the dimension of the latent space. We set the patch size as 16$\times$16 pixels, following \cite{Dosovitskiy2021}.

\item \paragraph{Patch-wise AE.}
We make further simplification on VAE by removing the regularization term:
\begin{equation}
\|x - U{^T}Vx\|^2. \nonumber
\end{equation}
As such, this tokenizer is essentially an autoencoder (AE) on patches, with the encoder and decoder both being linear projections.

\item \paragraph{Patch-wise PCA.}
Finally, we consider a simpler variant which performs Principal Component Analysis (PCA) on the patch space. It is easy to show that PCA is equivalent to a special case of AE:
\begin{equation}
\|x - V^TVx\|^2. \nonumber
\end{equation}
in which $V$ satisfies $VV^T{=}I$ (${d}{\times}{d}$ identity matrix).
The PCA bases can be simply computed by eigen-decomposition on a large set of randomly sampled patches, requiring no gradient-based training.
\end{itemize}

\vspace{1em}

\begin{figure}[t]
\centering
\begin{subfigure}[t]{\linewidth}
    \centering
    \includegraphics[width=.999\linewidth]{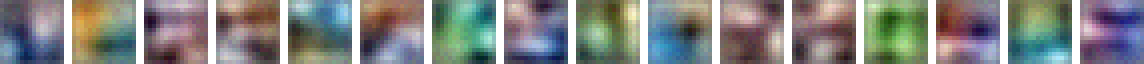}
    \caption{patch-wise VAE}
\vspace{.4em}
\end{subfigure}
\begin{subfigure}[t]{\linewidth}
    \centering
    \includegraphics[width=.999\linewidth]{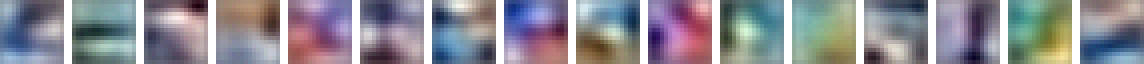}
    \caption{patch-wise AE}
\vspace{.4em}
\end{subfigure}
\begin{subfigure}[t]{\linewidth}
    \centering
    \includegraphics[width=.999\linewidth]{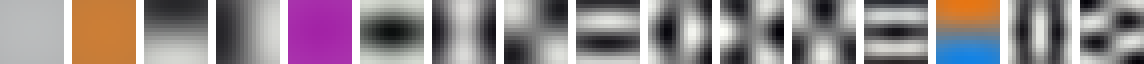}
    \caption{patch-wise PCA}
\end{subfigure}
\vspace{-.5em}
\caption{\label{fig:pca_bases}\textbf{Visualization of the patch-wise tokenizer}. Each filter corresponds to a row of the linear projection matrix $V$ ($d{\times}D$), reshaped to 16$\times$16$\times$3 for visualization. Here $d{=}16$.
}
\end{figure}

\noindent Thanks to the simplicity of using patches, for the three patch-wise tokenizers, we can visualize their filters in the patch space (\cref{fig:pca_bases}).

\cref{fig:lat_dim} summarizes the linear probe accuracy of DiT using these four variants of tokenizers. We show the results \wrt the \textit{latent dimension ``per token"}.\footnotemark 
~We draw the following observations.

\footnotetext{
For patch-wise VAE/AE/PCA (patch stride is 16), we treat each patch as a token, so the latent dimension is simply $d$ for each patch.
For the default convolutional VAE that has a stride of $8$, the DiT implementation \cite{Peebles2023} treats each 2$\times$2 patch on the latent space as a ``token"; as a result, its latent dimension ``per token" should be multiplied by 4 for calibration.
}

\begin{table}[t]
\centering
\vspace{-1em}
\includegraphics[width=.9\linewidth]{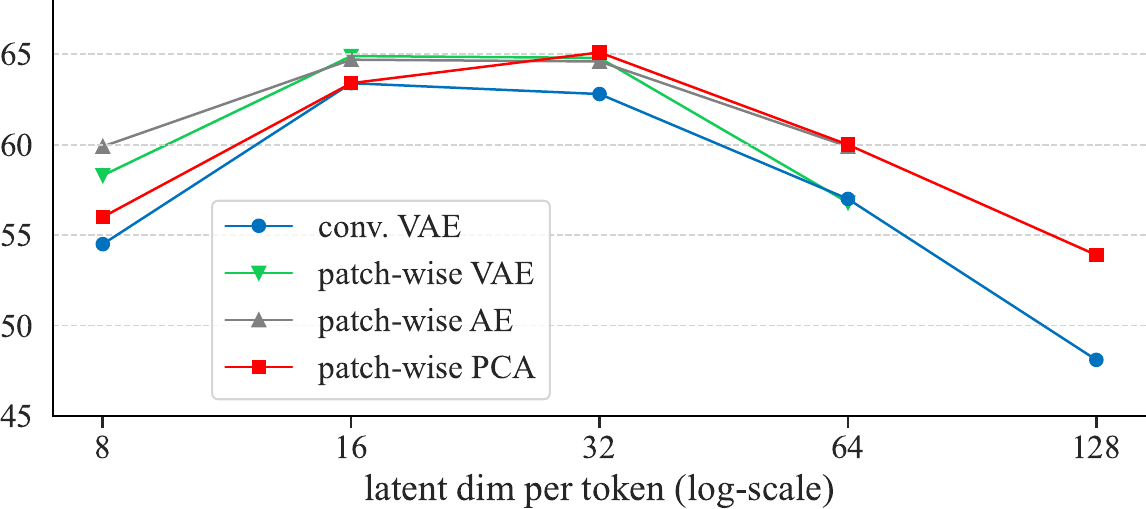}
\vspace{-.5em}
\tablestyle{6pt}{1.2}
\begin{center}
{\scriptsize
    \begin{tabular}{y{60}x{20}x{20}x{20}x{20}x{20}}
    latent dim. $d$ & 8 & 16 & 32 & 64 & 128 \\
    \hshline
    conv. VAE (baseline) & 54.5 & \textbf{63.4} & 62.8 & 57.0 & 48.1 \\
    patch-wise VAE & 58.3 & \textbf{64.9} & 64.8 & 56.8 & - \\
    patch-wise AE & 59.9 & \textbf{64.7} & 64.6 & 59.9 & - \\
    patch-wise PCA & 56.0 & 63.4 & \textbf{65.1} & 60.0 & 53.9 \\
    \end{tabular}
}
\end{center}
\vspace{-1em}
\caption{\label{fig:lat_dim}\textbf{Linear probe accuracy \vs latent dimension}. With a DiT model, we study four variants of tokenizers for computing the latent space.
We vary the dimensionality $d$ (\textit{per token}) of the latent space.
The table is visualized by the plot above.
\textbf{All four variants of tokenizers exhibit similar trends}, despite their differences in architectures and loss functions.
The 63.4\% entry of ``conv. VAE" is the same entry as the last line in \cref{tab:dit_ssl}.
}
\end{table}

\subsubsection*{\textit{Latent dimension of the tokenizer is crucial for DDM to work well in self-supervised learning.}}

As shown in \cref{fig:lat_dim}, all four variants of tokenizers exhibit similar trends, despite their differences in architectures and loss functions.
Interestingly, the optimal dimension is relatively low ($d$ is 16 or 32), even though the full dimension per patch is much higher (16$\times$16$\times$3=768).

Surprisingly, the convolutional~VAE tokenizer is neither necessary nor favorable; instead, all patch-based tokenizers, in which each patch is encoded \textit{independently}, perform similarly with each other and consistently outperform the Conv VAE variant. In addition, the KL regularization term is \textit{unnecessary}, as both the AE and PCA variants work well.

To our further surprise, \textit{even the PCA tokenizer works well.} Unlike the VAE or AE counterparts, the PCA tokenizer does \textit{not} require gradient-based training. With pre-computed PCA bases, the application of the PCA tokenizer is analogous to a form of image pre-processing, rather than a ``network architecture". The effectiveness of a PCA tokenizer largely helps us push the modern DDM towards a classical DAE, as we will show in the next subsection.

\begin{figure}[t]
\centering
\includegraphics[width=.9\linewidth]{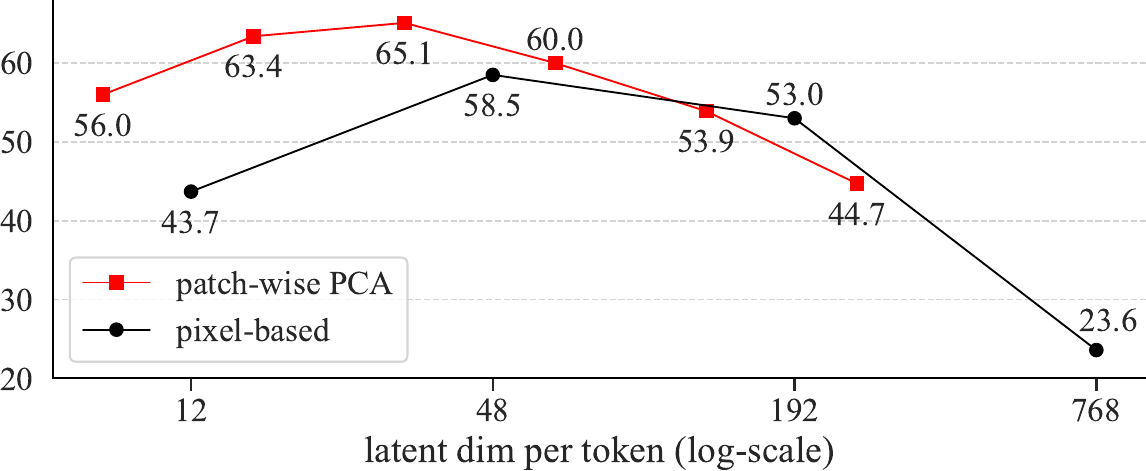}
\vspace{-.5em}
\caption{\label{fig:pca_pixel}
Linear probe results of the \textbf{pixel-based tokenizer}, operated on an image size of 256, 128, 64, and 32, respectively with a patch size of 16, 8, 4, 2. The ``latent" dimensions of these tokenized spaces are 768, 192, 48, and 12 per token.
Similar to other tokenizers we study, this pixel-based tokenizer exhibits a similar trend: a relatively small dimension of the latent space is optimal.
}
\vspace{-.5em}
\end{figure}

\subsubsection*{\textit{High-resolution, pixel-based DDMs are inferior for self-supervised learning.}}

Before we move on, we report an extra ablation that is consistent with the aforementioned observation. Specifically, we consider a ``na\"{i}ve tokenizer" that performs \textit{identity mapping} on patches extracted from resized images. In this case, a ``token" is the flatten vector consisting all \textit{pixels} of a patch.  

In \cref{fig:pca_pixel}, we show the results of this ``pixel-based" tokenizer, operated on an image size of 256, 128, 64, and 32, respectively with a patch size of 16, 8, 4, 2. The ``latent" dimensions of these tokenized spaces are 768, 192, 48, and 12 per token. In all case, the sequence length of the Transformer is kept unchanged (256).

Interestingly, this \textit{pixel-based} tokenizer exhibits a similar trend with other tokenizers we have studied, although the optimal dimension is shifted. In particular, the optimal dimension is $d$=48, which corresponds to an image size of 64 with a patch size of 4.
With an image size of 256 and a patch size of 16 ($d$=768), the linear probe accuracy drops dramatically to 23.6\%.

These comparisons show that the tokenizer and the resulting latent space are crucial for DDM/DAE to work competitively in the self-supervised learning scenario. In particular, applying a classical DAE with additive Gaussian noise on the pixel space leads to poor results.

\subsection{Toward Classical Denoising Autoencoders \label{sec:to_pixel}}

Next, we go on with our deconstruction trajectory and aim to get as close as possible to the classical DAE~\cite{Vincent2008}. We attempt to remove every single aspect that still remains between our current PCA-based DDM and the classical DAE practice.
Via this deconstructive process, we gain better understandings on how every modern design may influence the classical DAE.
\cref{tab:to_pixel} gives the results, discussed next.

\begin{table}[t]
\vspace{-1em}
\centering
\tablestyle{7pt}{1.1}
\begin{tabular}{y{160}x{50}}
 & acc. \\
\shline
patch-wise PCA baseline & 65.1\\
+ predict clean data (rather than noise) & 62.4 \\
~~ + remove input scaling (fix $\alphat\equiv1$) & 63.6 \\
\hline
~~~~ + operate on image input with inv. PCA & 63.6 \\
~~~~~~ + operate on image output with inv. PCA & 63.9 \\
~~~~~~~~ + predict original image & 64.5 \\
\end{tabular}
\vspace{-0.7em}
\caption{\label{tab:to_pixel}\textbf{Moving toward a classical DAE}, starting from our patch-wise PCA tokenizer.
Each line is based on a modification of the immediately preceding line.
See \cref{sec:to_pixel} for descriptions.
}
\vspace{-0.5em}
\end{table}

\paragraph{Predict clean data (rather than noise).}
While modern DDMs commonly predict the noise $\epst$ (see \cref{eq:loss}), the classical DAE predicts the clean data instead. We examine this difference by minimizing the following loss function:
\begin{equation}\label{eq:loss_clean}
 \lambda_t   \|\x0 - \mathtt{net}(z_t)\|^2
\end{equation}
Here $\x0$ is the clean data (in the latent space), and $\mathtt{net}(z_t)$ is the network prediction. $\lambda_t$ is a $t$-dependent loss weight, introduced to balance the contribution of different noise levels~\cite{Salimans2022}. It is suggested to set $\lambda_t={\alphat^2}/{\sigmat^2}$ as per~\cite{Salimans2022}. 
We find that setting $\lambda_t = {\alphat^2}$ works better in our scenario. Intuitively, it simply puts more weight to the loss terms of the \textit{cleaner} data (larger $\alphat$).

With the modification of predicting clean data (rather than noise), the linear probe accuracy \textit{degrades} from 65.1\% to 62.4\% (\cref{tab:to_pixel}). This suggests that the choice of the prediction target influences the representation quality.

Even though we suffer from a degradation in this step, we will stick to this modification from now on, as our goal is to move towards a classical DAE.\footnotemark

\footnotetext{We have revisited undoing this change in our final entry, in which we have not observed this degradation.}

\paragraph{Remove input scaling.} 
In modern DDMs (see \cref{eq:diffuse}), the input is scaled by a factor of $\alphat$.
This is not common practice in a classical DAE. Next, we study removing input scaling, \ie, we set $\alphat\equiv1$.
As $\alphat$ is fixed, we need to define a noise schedule directly on $\sigmat$. We simply set $\sigmat$ as a linear schedule from 0 to $\sqrt{2}$. Moreover, we empirically set the weight in \cref{eq:loss_clean} as $\lambda_t={1}/{(1+\sigmat^2)}$, which again puts more emphasis on cleaner data (smaller $\sigmat$).

After fixing $\alphat\equiv1$, we achieve a decent accuracy of 63.6\% (\cref{tab:to_pixel}), which compares favorably with the varying $\alphat$ counterpart's 62.4\%. This suggests that scaling the data by $\alphat$ is not necessary in our scenario.

\paragraph{Operate on the \textit{image} space with inverse PCA.} Thus far, for all entries we have explored (except \cref{fig:pca_pixel}), the model operates on the latent space produced by a tokenizer (\cref{fig:dit_arch} (b)). Ideally, we hope our DAE can work directly on the \textit{image} space while still having good accuracy. With the usage of PCA, we can achieve this goal by inverse PCA.

The idea is illustrated in \cref{fig:teaser}. Specially, we project the input image into the latent space by the PCA bases (\ie, $V$), add noise in the latent space, and project the noisy latent back to the image space by the \textit{inverse} PCA bases ($V^T$). \cref{fig:teaser} (middle, bottom) shows an example image with noise added in the latent space. With this noisy image as the input to the network, we can apply a standard ViT network \cite{Dosovitskiy2021} that directly operate on images, as if there is no tokenizer.

Applying this modification on the input side (while still predicting the output on the latent space) has 63.6\% accuracy (\cref{tab:to_pixel}). Further applying it to the output side (\ie, predicting the output on the image space with inverse PCA) has 63.9\% accuracy. Both results show that operating on the image space with inverse PCA can achieve similar results as operating on the latent space.

\paragraph{Predict original image.} While inverse PCA can produce a prediction target in the image space, this target is not the original image. This is because PCA is a \textit{lossy} encoder for any reduced dimension $d$. In contrast, it is a more natural solution to predict the original image directly.

When we let the network predict the original image, the ``noise'' introduced includes two parts: (i) the additive Gaussian noise, whose intrinsic dimension is $d$, and (ii) the PCA reconstruction error, whose intrinsic dimension is \mbox{$D-d$} ($D$ is 768). We weight the loss of both parts differently.

Formally, with the clean original image $x_0$ and network prediction $\mathtt{net}(x_t)$, we can compute the residue $r$ projected onto the full PCA space: \mbox{$r \triangleq V(x_0-\mathtt{net}(x_t))$}. Here $V$ is the $D$-by-$D$ matrix representing the full PCA bases.
Then we minimize the following loss function:
\begin{equation}\label{eq:overall}
    \lambda_t \sum_{i=1}^D w_i {r_i^2}.
\end{equation}
Here $i$ denotes the $i$-th dimension of the vector $r$. The per-dimension weight $w_i$ is 1 for $i \leq d$, and 0.1 for $d < i \leq D$.
Intuitively, $w_i$ down-weights the loss of the PCA reconstruction error.
With this formulation, predicting the original image achieves 64.5\% linear probe accuracy (\cref{tab:to_pixel}).

This variant is conceptually very simple: its input is a noisy image whose noise is added in the PCA latent space, its prediction is the original clean image (\cref{fig:teaser}).

\paragraph{Single noise level.} 
Lastly, out of curiosity, we further study a variant with \textit{single-level} noise.
We note that multi-level noise, given by noise scheduling, is a property motived by the diffusion process in DDMs; it is conceptually unnecessary in a classical DAE. 

We fix the noise level $\sigma$ as a constant ($\sqrt{1/3}$).
Using this single-level noise achieves {decent} accuracy of 61.5\%, a 3\% degradation \vs the multi-level noise counterpart (64.5\%). 
Using multiple levels of noise is analogous to a form of data augmentation in DAE: it is beneficial, but not an enabling factor. This also implies that the representation capability of DDM is mainly gained by the denoising-driven process, not a diffusion-driven process.

As multi-level noise is useful and conceptually simple, we keep it in our final entries presented in the next section. 

\begin{figure}[t]
\centering
\vspace{-1em}
\includegraphics[width=.9\linewidth]{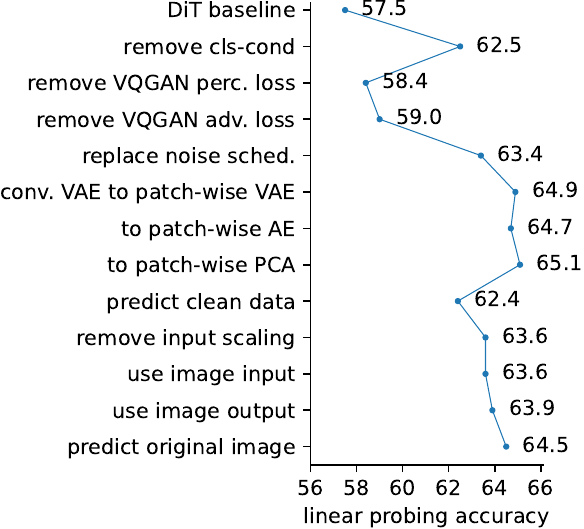}
\vspace{-.5em}
\caption{\label{fig:overall}\textbf{The overall deconstructive trajectory} from a modern DDM to \ours, summarizing \cref{tab:dit_ssl}, \cref{fig:lat_dim}, and \cref{tab:to_pixel}.
Each line is based on a modification of the immediately preceding line.
}
\vspace{-.5em}
\end{figure}

\subsection{Summary \label{sec:summary}}

In sum, we deconstruct a modern DDM and push it towards a classical DAE (\cref{fig:overall}).
We \textit{undo} many of the modern designs and conceptually retain only two designs inherited from modern DDMs: (i) a low-dimensional latent space in which noise is added; and (ii) multi-level noise. 

We use the entry at the end of \cref{tab:to_pixel} as our final DAE instantiation (illustrated in \cref{fig:teaser}).
We refer to this method as ``\emph{latent} Denoising Autoencoder'', or in short, \textbf{\ours}.

\section{Analysis and Comparisons\label{sec:exper}}


\begin{figure}[t]
\centering
\includegraphics[width=.99\linewidth]{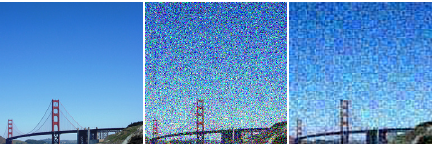}
\vspace{-.5em}
\caption{\label{fig:noise}
\textbf{Visualization: pixel noise \vs latent noise}.
\textbf{Left}: clean image, 256$\times$256 pixels. \textbf{Middle}: Gaussian noise added to the pixel space. \textbf{Right}: Gaussian noise added to the latent space produced by the PCA tokenizer, visualized by back projection to the image space using inverse PCA. $\sigma{=}\sqrt{1/3}$ in both cases.
}
\vspace{-.5em}
\end{figure}

\begin{figure*}[t]
\vspace{-2em}
\centering
\includegraphics[width=.99\linewidth]{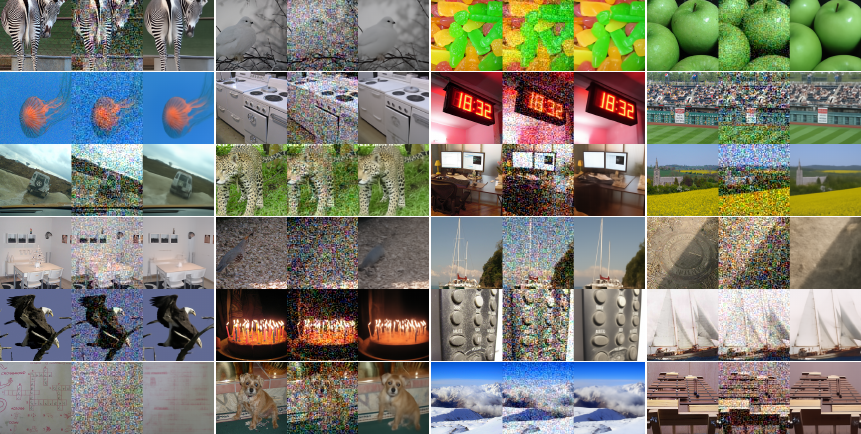}
\vspace{-.5em}
\caption{\label{fig:vis}\textbf{Denoising results} of \ours, evaluated on ImageNet validation images. 
\textit{This denoising problem, serving as a pretext task, encourages the network to learn meaningful representations in a self-supervised manner.}
For each case, we show: (\textbf{left}) clean image; (\textbf{middle}) noisy image that is the input to the network, where the noise is added to the latent space; (\textbf{right}) denoised output.
}
 \vspace{-1em}
\end{figure*}


\paragraph{Visualizing latent noise.} 
Conceptually, \ours is a form of DAE that learns to remove noise added to the latent space.
Thanks to the simplicity of PCA, we can easily visualize the latent noise by inverse PCA.

\cref{fig:noise} compares the noise added to pixels \vs to the latent. Unlike the pixel noise, the latent noise is largely independent of the \textit{resolution} of the image. With patch-wise PCA as the tokenizer, the pattern of the latent noise is mainly determined by the patch size. Intuitively, we may think of it as using patches, rather than pixels, to resolve the image. This behavior resembles MAE \cite{He2022}, which masks out patches instead of individual pixels.

\paragraph{Denoising results.} \cref{fig:vis} shows more examples of denoising results based on \ours. Our method produces reasonable predictions despite of the heavy noise. We note that this is less of a surprise, because neural network-based image restoration \cite{Burger2012,Dong2014} has been an intensively studied field.

Nevertheless, the visualization may help us better understand how \ours may learn good representations. The heavy noise added to the latent space creates a challenging \textit{pretext} task for the model to solve. It is nontrivial (even for human beings) to predict the content based on one or a few noisy patches locally; the model is forced to learn higher-level, more holistic semantics to make sense of the underlying objects and scenes.



\paragraph{Data augmentation.} 
Notably, all models we present thus far have \textit{no data augmentation}: only the center crops of images are used, with no random resizing or color jittering, following \cite{Dhariwal2021,Peebles2023}.
We further explore a mild data augmentation (random resized crop) for our final \ours:
\begin{center}\vspace{-.2em}
\tablestyle{4pt}{1.1}
\begin{tabular}{x{40}|x{48}x{48}}
aug. & center crop & random crop \\
\shline
acc. & 64.5 & 65.0
\end{tabular}\vspace{-.2em}
\end{center}
\noindent which has slight improvement. This suggests that \textit{the representation learning ability of $l$-DAE is largely independent of its reliance on data augmentation.} A similar behavior was observed in MAE \cite{He2022}, which sharply differs from the behavior of contrastive learning methods (\eg, \cite{Chen2020}).

\paragraph{Training epochs.}
All our experiments thus far are based on 400-epoch training.
Following MAE~\cite{He2022}, we also study training for 800 and 1600 epochs:
\begin{center}\vspace{-.2em}
\tablestyle{4pt}{1.1}
\begin{tabular}{x{40}|x{40}x{40}x{40}}
epochs & 400 & 800 & 1600 \\
\shline
acc. & 65.0 & 67.5 & 69.6
\end{tabular}\vspace{-.2em}
\end{center}
As a reference, MAE~\cite{He2022} has a significant gain (4\%) extending from 400 to 800 epochs, and MoCo v3~\cite{Chen2021a} has nearly no gain (0.2\%) extending from 300 to 600 epochs.

\paragraph{Model size.}
Thus far, our all models are based on the DiT-L variant \cite{Peebles2023}, whose encoder and decoder are both ``ViT-$\frac{1}{2}$L" (half depth of ViT-L).
We further train models of different sizes, whose encoder is ViT-B or ViT-L (decoder is always of the same size as encoder):
\begin{center}\vspace{-.2em}
\tablestyle{4pt}{1.1}
\begin{tabular}{x{40}|x{40}x{40}x{40}}
encoder & ViT-B & ViT-$\frac{1}{2}$L & ViT-L \\ [2pt]
\shline
acc. & 60.3 & 65.0 & 70.9
\end{tabular}\vspace{-.2em}
\end{center}
We observe a good \textit{scaling} behavior \wrt model size: scaling from ViT-B to ViT-L has a large gain of 10.6\%. A similar scaling behavior is also observed in MAE \cite{He2022}, which has a 7.8\% gain from ViT-B to ViT-L.

\paragraph{Comparison with previous baselines.} Finally, to have a better sense of how different families of self-supervised learning methods perform, we compare with previous baselines in \cref{tab:system}. We consider MoCo v3~\cite{Chen2021a}, which belongs to the family of contrastive learning methods, and MAE~\cite{He2022}, which belongs to the family of masking-based methods.

Interestingly, \ours performs decently in comparison with MAE, showing a degradation of 1.4\% (ViT-B) or 0.8\% (ViT-L). We note that here the training settings are made \textit{as fair as possible} between MAE and \ours: both are trained for 1600 epochs and with random crop as the data augmentation. On the other hand, we should also note that MAE is more efficient in training because it only operates on unmasked patches. Nevertheless, \textit{we have largely closed the accuracy gap between MAE and a DAE-driven method}.

Last, we observe that \textit{autoencoder-based} methods (MAE and \ours) still fall short in comparison with contrastive learning methods under this protocol, especially when the model is small. We hope our study will draw more attention to the research on autoencoder-based methods for self-supervised learning. 

\section{Conclusion}

We have reported that \ours, which largely resembles the classical DAE, can perform competitively in self-supervised learning.
The critical component is a low-dimensional latent space on which noise is added.
We hope our discovery will reignite interest in denoising-based methods in the context of today's self-supervised learning research. 

\begin{table}[t]
\tablestyle{8pt}{1.05}
\begin{tabular}{x{40}|cc}
method &
ViT-B (86M)
&
ViT-L (304M)
\\
\shline
MoCo v3 & 76.7 & 77.6 \\
MAE & 68.0 & 75.8 \\
\ours & 66.6 & 75.0 \\
\end{tabular}

\vspace{-1em}
\caption{\label{tab:system}\textbf{Comparisons with previous baselines} of MoCo v3~\cite{Chen2021a} and MAE~\cite{He2022}. 
The entries of both MAE and \mbox{\ours} are trained for 1600 epochs and with random crop as the data augmentation. Linear probe accuracy on ImageNet is reported. In the brackets are the number of parameters of the encoder.
}
 \vspace{-.5em}
\end{table}

\clearpage

\paragraph{Acknowledgement.} We thank Pascal Vincent, Mike Rabbat, and Ross Girshick
 for their discussion and feedback.

\setcounter{section}{0}
\renewcommand\thesection{\Alph{section}}

\section{Implementation Details\label{sec:impl}}

\paragraph{DiT architecture.} 
We follow the DiT architecture design~\cite{Peebles2023}.
The DiT architecture is similar to the original ViT~\cite{Dosovitskiy2021}, with extra modifications made for conditioning.
Each Transformer block accepts an embedding network (a two-layer MLP) conditioned on the time step $t$. The output of this embedding network determines the scale and bias parameters of LayerNorm \cite{Ba2016}, referred to as adaLN~\cite{Peebles2023}.
Slightly different from~\cite{Peebles2023}, we set the hidden dimension of this MLP as 1/4 of its original dimension, which helps reduce model sizes and save memory, at no accuracy cost.
 
\paragraph{Training.} The original DiTs \cite{Peebles2023} are trained with a batch size of 256. To speed up our exploration, we increase the batch size to 2048. We perform linear learning rate warm up \cite{Goyal2017} for 100 epochs and then decay it following a half-cycle cosine schedule.
We use a base learning rate \textit{blr} = 1e-4 \cite{Peebles2023} by default, and set the actual \textit{lr} following the linear scaling rule \cite{Goyal2017}: \textit{blr} $\times$ batch\_size / 256.
No weight decay is used \cite{Peebles2023}. 
We train for 400 epochs by default.
On a 256-core TPU-v3 pod, training DiT-L takes 12 hours.

\paragraph{Linear probing.}
Our linear probing implementation follows the practice of MAE \cite{He2022}.
We use \textit{clean}, $256{\times}256$-sized images for linear probing training and evaluation. The ViT output feature map is globally pooled by average pooling. It is then processed by a parameter-free BatchNorm~\cite{Ioffe2015} layer and a linear classifier layer, following \cite{He2022}.
The training batch size is 16384, learning rate is $6.4{\times}10^{-3}$ (cosine decay schedule), weight decay is 0, and training length is 90 epochs. Randomly resized crop and flipping are used during training and a single center crop is used for testing. Top-1 accuracy is reported.

While the model is conditioned on $t$ in self-supervised pre-training, conditioning is not needed in transfer learning (\eg, linear probing).
We fix the time step $t$ value in our linear probing training and evaluation.
The influence of different $t$ values (out of 1000 time steps) is shown as follows:
\begin{center}\vspace{-.2em}
\tablestyle{4pt}{1.1}
\begin{tabular}{x{64}|x{20}x{20}x{20}x{20}x{20}}
fixed $t$ & 0 & 10 & 20 & 40 & 80 \\
\shline
w/ clean input & 64.1 & 64.5 & 64.1 & 63.3 & 62.2 \\
w/ noisy input & 64.2 & 65.0 & 65.0 & 65.0 & 64.5
\end{tabular}\vspace{-.2em}
\end{center}
\noindent We note that the $t$ value determines: (i) the model weights, which are conditioned on $t$, and (ii) the noise added in transfer learning, using the same level of $t$. Both are shown in this table. We use $t$ = 10 and clean input in all our experiments, except \cref{tab:system} where we use the optimal setting.

Fixing $t$ also means that the $t$-dependent MLP layers, which are used for conditioning, are not exposed in transfer learning, because they can be merged given the fixed $t$. As such, our model has the number of parameters just similar to the standard ViT \cite{Dosovitskiy2021}, as reported in \cref{tab:system}.

The DiT-L~\cite{Peebles2023} has 24 blocks where the first 12 blocks are referred to as the ``encoder" (hence \mbox{ViT-$\frac{1}{2}$L}) and the others the ``decoder". This separation of the encoder and decoder is artificial. In the following table, we show the linear probing results using different numbers of blocks in the encoder, using the same pre-trained model:
\begin{center}\vspace{-.2em}
\tablestyle{4pt}{1.1}
\begin{tabular}{c|x{18}x{18}x{18}x{18}x{18}x{18}x{18}}
enc. blocks & 9 & 10 & 11 & 12 & 13 & 14 & 15 \\
\shline
acc. & 58.5 & 62.0 & 64.1 & \textbf{64.5} & 63.6 & 61.9 & 59.7
\end{tabular}\vspace{-.2em}
\end{center}
The optimal accuracy is achieved when the encoder and decoder have the same depth. This behavior is different from MAE's \cite{He2022}, whose encoder and decoder are asymmetric.

\section{Fine-tuning Results\label{sec:ft}}

In addition to linear probing, we also report end-to-end fine-tuning results. 
We closely followed MAE's protocol~\cite{He2022}.
We use clean, $256{\times}256$-sized images as the inputs to the encoder.
Globally average pooled outputs are used as features for classification.
The training batch size is 1024, initial learning rate is $4{\times}10^{-3}$, weight decay is 0.05, drop path~\cite{Huang2016} is 0.1, and training length is 100 epochs. 
We use a layer-wise learning rate decay of 0.85 (B) or 0.65 (L).
MixUp~\cite{Zhang2018a} (0.8), CutMix~\cite{Yun2019} (1.0), RandAug~\cite{Cubuk2020} (9, 0.5), and exponential moving average (0.9999) are used, similar to~\cite{He2022}.
The results are summarized as below:
\begin{center}\vspace{-.3em}
\tablestyle{8pt}{1.05}
\begin{tabular}{x{40}|cc}
method & ViT-B & ViT-L \\
\shline
MoCo v3 & 83.2 & 84.1 \\
MAE & 83.6 & 85.9 \\
\ours & 83.7 & 84.7 \\
\end{tabular}
\vspace{-.3em}
\end{center}
\noindent
Overall, both autoencoder-based methods are better than MoCo v3.
\ours performs similarly with MAE with \mbox{ViT-B}, but still fall short of MAE with ViT-L.

{
\small
\bibliographystyle{ieeenat_fullname}
\bibliography{dae}
}

\end{document}